\documentclass[11pt]{article}

\usepackage[T1]{fontenc}
\usepackage[utf8]{inputenc}
\usepackage{lmodern}
\usepackage[margin=1in]{geometry}
\usepackage{amsmath,amssymb}
\usepackage{graphicx}
\usepackage{booktabs}
\usepackage{multirow}
\usepackage[round,authoryear]{natbib}
\usepackage[hidelinks]{hyperref}
\usepackage[capitalise,noabbrev]{cleveref}
\usepackage{xcolor}

\newcommand{\pp}{\,pp}  
\newcommand{\codeurl}{\url{https://github.com/Effiani-Core/memory-or-clock}}
\newcommand{\dataurl}{\url{https://doi.org/10.5281/zenodo.23512311}}

\title{Memory or a Clock?\\
\large Auditing a Memory Advantage in XLand-MiniGrid with Seeds, Maps and Evaluation Noise Apart}

\author{Rodrigo Reveron-Fonseca\\
Tecnol\'ogico de Monterrey\\
\texttt{a01841051@tec.mx}}

\date{}

\begin{document}
\maketitle

\begin{abstract}
On a partially observable XLand-MiniGrid task, an LSTM agent beats a memory-free MLP by 25
percentage points. The gap looks like a clear case of memory. We ask two questions about it.
First, how much of a measured gap is noise? Separating training seed, map and evaluation
randomness in 60 runs, each re-evaluated three times, we find that about ten paired seeds detect
a 10-point gap in final success. The binary ``task acquired'' verdict, a common summary of
learning curves, is 76\% evaluation noise and would need about a hundred seeds for the same gap.
Second, is the gap memory? On a sealed panel of new maps, opened once, a memory-free MLP of the
LSTM's size comes within 3.3 points of the LSTM when it is given the time step, and gains no
measurable benefit from its width without it. With the time step, an MLP a quarter of the
LSTM's size, and the LSTM with its state erased at every step, also come within about 5 points.
A linear probe reads the time step from the trained LSTM's state. On this task, most of the
apparent memory advantage is an advantage of knowing the time, which matters because episodes
are cut at a fixed limit and the reward shrinks with time. We recommend giving memory-free baselines the time step,
and we release the protocol, the event-driven tooling and every checkpoint.
\end{abstract}

\section{Introduction}
\label{sec:intro}

When an agent cannot see everything that matters, it has to remember. The standard test of
whether memory helps is a comparison: train a recurrent agent and a feedforward one on a partially
observable task and see which does better. If the recurrent agent wins, the gap is read as the
value of memory. Small, fast environments such as XLand-MiniGrid \citep{nikulin2024xland} make
such comparisons cheap enough to run on a laptop, and so they are run often.

We started from one such comparison. In a pilot with three seeds on an XLand-MiniGrid pickup task,
an LSTM agent learned the task and a memory-free MLP did not; on held-out maps the LSTM was ahead
by 17 to 42 percentage points on every seed. It looked like a clean memory result. Before building
on it we asked two questions that apply to any result of this kind.

\textbf{How much of a measured gap is noise?} A run's score depends on the training seed, on the
maps it is tested on, and on the randomness of the evaluation episodes themselves. With three
seeds and a single evaluation per checkpoint these sources are mixed, and some popular summaries,
such as a yes/no verdict that a task was ``acquired'', may be dominated by the last one. Prior
work has shown that deep RL comparisons need more seeds than are usually run
\citep{henderson2018matters,colas2018seeds,agarwal2021precipice,patterson2024empirical}; we ask
how many, for which metric, and why, by separating the three sources of variation with the
tools of generalizability theory \citep{cronbach1972dependability,brennan2001generalizability}.

\textbf{Is the gap memory?} The LSTM in our pilot was also four times larger than the MLP, and
it could do something the MLP could not: count the steps of the episode. Episodes are cut at a
fixed time limit and the reward shrinks with time, so knowing the time is useful on its own
\citep{pardo2018timelimits}. A memory advantage is only an advantage of memory once capacity and
time awareness are controlled.

We answer both questions with 188 runs on one laptop CPU, every one listed in a plan before its
results existed, executed as an event-driven chain and re-evaluated from saved weights. Our
contributions are:
\begin{itemize}
  \item \textbf{A variance decomposition of an RL comparison} into training seed, map, seed
  $\times$ map and evaluation noise (60 runs, \cref{sec:results}), with the number of seeds each
  metric needs. About ten paired seeds detect a 10-point gap in final success, but the binary
  acquisition verdict is 76\% evaluation noise and needs about a hundred.
  \item \textbf{An audit of the memory advantage} with parameter-matched controls, a
  recurrence ablation, a short fixed memory and a clock input, tested on a sealed panel of new
  maps opened once (\cref{sec:part2}). A memory-free MLP of the LSTM's size that is told the time
  step comes within 3.3 points of the LSTM; width without the clock adds nothing measurable.
  Most of the apparent memory advantage is an advantage of knowing the time.
  \item \textbf{Evidence for the mechanism}: a linear probe reads the time step from the trained
  LSTM's state, but not from the observation or from an LSTM whose state is erased; and giving
  the time step to that erased LSTM, or to an MLP a quarter of the LSTM's size, brings either to
  within 5 points of the LSTM.
  \item \textbf{Tooling and data}: preregistered plans, an event-driven runner with no silent
  retries, re-evaluation of every checkpoint, sealed panels, and all weights and per-episode
  outcomes. Re-running a seed reproduces its weights bit for bit.
\end{itemize}

\section{Related work}
\label{sec:related}

\paragraph{Statistical practice in deep RL.}
\citet{henderson2018matters} showed that results in deep RL can change with the random seed and
with implementation details, and \citet{colas2018seeds} turned this into power analyses for the
number of seeds. \citet{jordan2020evaluating} proposed performance measures that aggregate
across environments and account for variation across runs, \citet{agarwal2021precipice} recommended interval
estimates and robust aggregates such as the interquartile mean when only a few runs are
affordable, and \citet{patterson2024empirical} gave broad guidance on empirical design. These
works treat the run as the unit and its score as given. We look inside the score: in a
procedurally generated environment a run's score also depends on which maps it is tested on and
on the sampling randomness of the evaluation episodes, and these sources call for different
remedies (more seeds, more maps, or more evaluations). We estimate them separately and show that
they differ by metric, to the point that one common summary, the acquisition verdict, is mostly
evaluation noise.

\paragraph{Generalizability theory.}
Separating the sources of variance of a measurement, and planning a design from them, is the
subject of generalizability theory \citep{cronbach1972dependability,brennan2001generalizability},
long used for test scores and ratings, with variance components estimated as in
\citet{searle1992variance}. To our knowledge it has rarely been applied to the evaluation of RL
agents; we use its standard crossed model with training seeds, maps and evaluation replicas as
facets.

\paragraph{Evaluating memory in RL.}
\citet{hausknecht2015drqn} compared a recurrent Q-network that reads one frame at a time with a
feedforward one that reads a stack of four, the classic short-memory control we also use. POPGym
\citep{morad2023popgym} found memory-free policies competitive on several of its partially
observable tasks, \citet{ni2023transformers} separated memory length from credit assignment, and
\citet{cherepanov2024memory} argued that conclusions about memory depend on the type of memory a
task requires and on the evaluation protocol. Our controls follow the same spirit for one concrete
claim: we match parameters, ablate recurrence while keeping the weights, give a short fixed
memory, and, decisively here, give memory-free agents the time step.

\paragraph{Time limits.}
\citet{pardo2018timelimits} showed that when episodes are cut by a time limit, an agent that is
not told the remaining time faces a partially observable problem, and that adding the time to the
observation helps. \citet{deverett2019interval} showed that recurrent agents trained on interval
timing tasks develop clock-like dynamics in their state. Our result is an instance of these
effects hiding inside a memory comparison on a task that is not about timing: part of what a
recurrent agent gains from its state is a clock.

\paragraph{XLand-MiniGrid.}
XLand-MiniGrid \citep{nikulin2024xland} implements MiniGrid-style worlds
\citep{chevalierboisvert2023minigrid} with objects, goals and rules in JAX and is used mainly for
meta-reinforcement learning. We use a single fixed task from it and do not study meta-learning;
our findings concern how a memory comparison on such a task should be run and read.

\section{Methods}
\label{sec:methods}

\subsection{Environment and task}
\label{sec:env}

We use XLand-MiniGrid~0.9.3 \citep{nikulin2024xland}, a JAX re-implementation of grid worlds
with objects and rules in the style of MiniGrid \citep{chevalierboisvert2023minigrid}, running on
CPU. Every episode takes place on a $7\times7$ grid built by the environment's procedural
generator from an integer \emph{map seed}; the seed fixes the walls and the positions of the
agent and the objects. The agent sees only an egocentric $5\times5$ crop in front of it: for
each cell, the tile type and its colour, scaled to $[0,1]$ and flattened to 50 inputs. It never
receives the goal, the rules or the global map. It has six discrete actions (move forward, turn
left, turn right, pick up, put down, toggle).

Task~B places two objects on the grid: a red ball, which is the goal, and a green key, which is a
distractor. The episode succeeds, and ends, when the agent holds the red ball; it is truncated
after 32 steps. The reward is the environment's default sparse success reward,
$1 - 0.9\,t/32$ at the successful step $t$ and $0$ otherwise. Because the view is partial, the
target is often out of sight, and a policy without memory can condition only on the current crop.
A random policy succeeds on 15.6\% of the validation maps and 14.1\% of the test maps.

\paragraph{Map panels.}
Maps are drawn from fixed, disjoint \emph{panels} of map seeds: 64 training maps (the only maps
seen during training, cycled episode by episode), 32 validation maps (learning curves and
checkpoint selection), 64 test maps (the reported score), plus a sealed final panel and a
small known panel that this paper does not use. Panels contain no duplicate layout within or
across panels: we hash each map's physical state (grid, agent position and direction, and the
contents of the agent's pocket) and remove repeats.

\subsection{Architectures}
\label{sec:arch}

Both agents share an encoder: a two-layer MLP with $\tanh$ activations, $50\to32\to32$. The
\textbf{MLP} agent reads its policy and value directly from the encoder and has no memory
(2,919 trainable parameters). The \textbf{LSTM} agent adds a single LSTM cell
\citep{hochreiter1997lstm} with 32 units on top of the encoder (11,367 parameters). Both use a
linear policy head over the six actions and a linear value head. The recurrent state is set to
zero at the start of every episode, in training and in evaluation. The two agents differ in size
by a factor of 3.9; \cref{sec:part2} controls for that difference.

\subsection{Training}
\label{sec:training}

Agents are trained with synchronous advantage actor-critic (A2C; \citealp{mnih2016a3c}) on a
single environment, with Adam \citep{kingma2015adam}, learning rate $10^{-3}$, rollouts of 16
steps, discount $\gamma=0.99$, value-loss coefficient $0.5$, entropy coefficient $0.01$ and
gradient-norm clipping at $0.5$. Gradients flow through the recurrent state within a rollout and
are cut at rollout boundaries. Following \citet{pardo2018timelimits}, the 32-step time limit is
treated as a truncation: the return bootstraps from the value of the last observation instead of
treating it as terminal. Each run lasts 200,000 environment interactions. These settings were
fixed by the study protocol before the pilot and are the same for both agents.

\subsection{Evaluation}
\label{sec:eval}

\paragraph{During training.}
Every 10,000 interactions (21 points per run, including step 0) the current weights are frozen,
saved as a checkpoint, and played once on each of the 32 validation maps. Actions are sampled from
the policy (not taken greedily); the sampling generator of each episode is seeded with its map
seed, so an evaluation is a deterministic function of the weights. This gives the learning curve.
At the end of training, the final weights are played once on each of the 64 test maps; this
\emph{final test success} is the primary score of a run.

\paragraph{Derived metrics.}
The \emph{validation AUC} is the trapezoidal area under the learning curve divided by the
training length, so it rewards learning early as well as learning well. The protocol's
\emph{acquisition rule} declares a task acquired when two consecutive validation points reach
at least $\max(0.75,\; s_0+0.2,\; s_{\text{rand}}+0.2)$, where $s_0$ is the success at step~0 and
$s_{\text{rand}}$ that of the random policy; the acquisition step is the step of the second such
point. A run therefore has a yes/no \emph{acquired} verdict.

\paragraph{Re-evaluation.}
To measure the randomness of evaluation itself, every saved checkpoint (all 21 curve points and
the final weights) is played again $R=3$ times with the weights fixed, using a new sampling seed
for each map and replica, $\text{map seed} + r\cdot 10{,}000{,}019$ for $r=1,2,3$ (replica~0 is
the original evaluation). We keep the outcome of every episode. Re-evaluation never changes the
run it reads; it verifies each checkpoint's SHA-256 digest before loading it.

\subsection{Seeds and preregistration}
\label{sec:seeds}

A run is identified by its architecture and its \emph{training seed}, which fixes the initial
weights, the order in which training maps are visited and the action-sampling stream during
training. Before any run started, a plan file fixed the study, the two architectures, the task,
the budget, $R$, and the seed rule: the three pilot seeds (7, 19, 31) followed by 27 seeds drawn
from NumPy's \texttt{SeedSequence(20261018)} modulo $10^6$, skipping repeats. Both architectures
are trained on the same 30 seeds, so every comparison is paired by seed. The plan, and a hash of
it, are recorded in the chain's journal; a changed plan cannot continue an existing chain.

\subsection{Separating seed, map and evaluation noise}
\label{sec:gtheory}

For one architecture and one metric, let $y_{smr}$ be the outcome of seed $s$ on map $m$ in
re-evaluation replica $r$ (for final test success, $y_{smr}\in\{0,1\}$; for the AUC, the area
under that map's curve in that replica). We use the crossed random-effects model of
generalizability theory \citep{cronbach1972dependability,brennan2001generalizability}
\begin{equation}
  y_{smr} = \mu + a_s + b_m + (ab)_{sm} + e_{smr},
  \label{eq:model}
\end{equation}
with independent zero-mean effects of variance $\sigma^2_{\text{seed}}$ (real differences between
training seeds), $\sigma^2_{\text{map}}$ (some maps are harder for everyone),
$\sigma^2_{\text{seed}\times\text{map}}$ (a seed finds some maps easier than other seeds do) and
$\sigma^2_{\text{noise}}$ (the same weights on the same map give a different outcome under another
sampling seed). With $S$ seeds, $M$ maps and $R$ replicas, the balanced ANOVA estimators
\citep{searle1992variance} are
\begin{align*}
  \hat\sigma^2_{\text{noise}} &= \mathrm{MS}_E, &
  \hat\sigma^2_{\text{seed}\times\text{map}} &= (\mathrm{MS}_{SM}-\mathrm{MS}_E)/R,\\
  \hat\sigma^2_{\text{seed}} &= (\mathrm{MS}_S-\mathrm{MS}_{SM})/(MR), &
  \hat\sigma^2_{\text{map}} &= (\mathrm{MS}_M-\mathrm{MS}_{SM})/(SR),
\end{align*}
where $\mathrm{MS}$ are the usual mean squares; negative estimates are reported as zero. The
acquisition verdict has one value per seed and replica, so for it we use the one-way version
($M=1$, map terms absent). Paired differences between architectures (LSTM minus MLP on the same
seed, map and sampling seed) are analysed with the same model. Intervals for the components are
percentile intervals from 1,000 bootstrap resamples of seeds \citep{efron1993bootstrap}.

\paragraph{From components to designs (D study).}
The variance of an architecture's mean score over $n_s$ seeds, $n_m$ maps and $n_r$ evaluations is
\begin{equation}
  \operatorname{Var}(\bar y) = \frac{\sigma^2_{\text{seed}}}{n_s}
    + \frac{\sigma^2_{\text{seed}\times\text{map}}}{n_s n_m}
    + \frac{\sigma^2_{\text{noise}}}{n_s n_m n_r}
    \;\Big[+\frac{\sigma^2_{\text{map}}}{n_m}\Big],
  \label{eq:dstudy}
\end{equation}
where the bracketed term applies only to claims about new maps; on a fixed panel shared by all
runs it is common to every arm and cancels in a comparison. From \cref{eq:dstudy} we report
(i)~the smallest $n_s$ whose 95\% interval half-width $1.96\sqrt{\operatorname{Var}(\bar y)}$ is
below a target, and (ii)~for a paired difference with per-seed variance
$V=\sigma^2_{\text{seed}}+\sigma^2_{\text{seed}\times\text{map}}/n_m+\sigma^2_{\text{noise}}/(n_m n_r)$,
the seeds a two-sided test at $\alpha=0.05$ needs to find a true gap $\delta$ with power 0.8,
$n_s=\lceil(z_{0.975}+z_{0.8})^2 V/\delta^2\rceil$ (normal approximation, minimum 2). Finally,
the \emph{noise share} of a metric is the fraction of the between-seed variance of a run's score
from a single evaluation ($n_m$ maps, $n_r=1$) that is due to evaluation noise,
$(\sigma^2_{\text{noise}}/n_m)\,/\,V$. All computations are in
\texttt{ucl study-variance}, which we checked on synthetic data with known components.

\subsection{Execution}
\label{sec:execution}

Runs execute one at a time as a chain of events: the end of a run starts its re-evaluation, and
the end of a re-evaluation starts the next run. Every event is appended to the chain's journal; a
failed or interrupted step stops the chain and is shown as such until it is repeated by hand with
the same seed. Two runs were stopped and repeated with the same seed, their partial outputs kept:
one by a planned restart of the machine, and one by the runner's resource guard, which halts a
run when available memory falls below 400\,MiB. Everything ran on
one laptop CPU (Intel Core i7-12700H, WSL2, PyTorch~2.11 with two threads, JAX~0.4.38 on CPU).
In Part~I a run took 6.4 minutes for the LSTM and 4.7 for the MLP, and its re-evaluation about
one minute; the 60 runs used 12M training interactions and about 6.5 hours. Part~II's 128 runs
(including the learning-rate search and the mechanism arms) used 38.4M interactions and about
24 hours.

\subsection{Part II design: memory, capacity or a clock?}
\label{sec:methods-part2}

Part~I compares an LSTM with an MLP that is 3.9 times smaller and cannot tell how far into the
episode it is. Part~II separates those differences with controls that change one thing at a time
(\cref{tab:arms}). All share the encoder of \cref{sec:arch} and the training and evaluation of
\cref{sec:training,sec:eval}, with a budget of 300,000 interactions.

\begin{itemize}
  \item \textbf{MLP-wide + clock} (11,399 parameters) replaces the LSTM cell with a memory-free
  MLP of the same size, $33\to128\to32$, whose extra input is the time inside the episode,
  $t/32$ ($t=0$ at the first step). Memory-free arms get the clock because, under a time limit,
  an agent that can count steps has information a memory-free agent lacks
  \citep{pardo2018timelimits}.
  \item \textbf{LSTM-reset} (11,367) is the LSTM with its state $(h,c)$ set to zero before every
  step, in training and evaluation: the same parameters and optimiser, no memory. Its 4,096
  recurrent weights receive no signal, so it controls for recurrence, not for capacity.
  \item \textbf{stack4 + clock} (11,381) sees the last four observations concatenated (zeros
  before the fourth step) and the clock: a short, fixed memory without learned recurrence
  \citep{hausknecht2015drqn}. Its body is $33\to55\to32$.
  \item \textbf{MLP-wide, no clock} (11,401) is MLP-wide without the clock input
  ($32\to130\to32$). It separates width from the clock. This arm was added after an interim look
  at 46 of the 80 phase-1 runs and is analysed as exploratory.
  \item \textbf{LSTM} and the small \textbf{MLP} are those of Part~I.
  \item Two \textbf{mechanism arms}, added after the confirmatory result and exploratory, know
  the time but have no memory: \textbf{LSTM-reset + clock} (11,495 parameters) feeds the clock to
  the cell of LSTM-reset, and \textbf{MLP + clock} (2,951) feeds it to the small MLP's encoder.
  They run on the same first 10 seeds with the protocol's learning rate, and their checkpoints
  were played on the same sealed panel under a separate evaluation record.
\end{itemize}

\paragraph{Learning rates.}
The new memory-free arms might need a different learning rate from the protocol's $10^{-3}$.
Before phase~1, MLP-wide + clock and stack4 + clock were each trained with $10^{-3}$,
$3\cdot10^{-4}$ and $3\cdot10^{-3}$ on three tuning seeds that are excluded from every
evaluation \citep{eimer2023hyperparameters}. A rule fixed in the plan chose the rate with the
highest mean validation AUC (ties to the protocol's rate) and wrote the choice to the chain's
journal before phase~1 started: $10^{-3}$ for MLP-wide (AUC 0.62 versus 0.58 and 0.12) and
$3\cdot10^{-4}$ for stack4 (0.42 versus 0.31 and 0.12). LSTM, LSTM-reset and the small MLP keep the
protocol's rate; MLP-wide without clock uses MLP-wide's.

\paragraph{Seeds.}
Twenty new training seeds come from \texttt{SeedSequence(20261020)}, excluding the pilot seeds,
the 30 seeds of Part~I and the tuning seeds. LSTM, MLP-wide + clock and LSTM-reset run on all 20;
the small MLP, stack4 + clock and MLP-wide without clock run on the first 10, so every arm is
paired with every other on the seeds they share. Within each seed, the order of the arms is a
fixed permutation drawn from the master seed. Every run is re-evaluated three times as in
\cref{sec:eval}. Phase~1 has 80 runs and the clock control 10.

\paragraph{A sealed panel of new maps.}
The confirmatory score is measured on 64 task-B maps that no run has seen and no study panel
contains. They were built without loading any model, by walking map seeds from 3,150,000 and
keeping maps whose physical identity differs from all 192 distinct maps of the study's panels.
For each run we play its validation-selected checkpoint (\texttt{best\_B}, chosen over the 300,000
interactions) on every sealed map with four new sampling seeds, 256 episodes per run. The panel
was opened once, after the analysis below was committed; the opening is recorded with the hash
of that analysis, and the tooling refuses to open it again with a different one.

\paragraph{Hypotheses and tests.}
The C01 design fixed, before any Part~II run, three principal contrasts and their thresholds:
\textbf{P1} (capacity), LSTM minus MLP-wide + clock is at least 15\pp; \textbf{P2} (recurrence),
LSTM minus LSTM-reset is at least 15\pp; and \textbf{P3}, a fully observable twin of the task,
which we did not run but still count in the family. The secondary contrast \textbf{H3} asks
whether a short fixed memory falls short of the two strong arms, that is, whether stack4 stays
below their average: $\text{LSTM} + \text{MLP-wide} - 2\,\text{stack4} > 0$. The unit of analysis is the training seed; maps are not replicates. For
each contrast we compute its value per seed on the seeds its arms share, an exact two-sided
sign-flip permutation test over all $2^n$ sign patterns, a BCa bootstrap 95\% interval with
10,000 resamples \citep{efron1993bootstrap} and the $t$ interval as a sensitivity check. Holm's
correction \citep{holm1979simple} uses $m=3$ for the principal family and $m=5$ for the secondary
one. A contrast is \emph{supported} when its Holm-adjusted $p<0.05$, its mean reaches the
threshold and the lower end of its interval is above zero. Contrasts added after the interim
look form an exploratory family that is reported without correction or verdict.

\paragraph{A probe for time.}
To see whether the LSTM's state carries the time step, we play each run's validation-selected
checkpoint on the 32 validation maps with two new sampling seeds and record, at every step $t$,
the LSTM's output after reading that step's observation. A ridge regression ($\lambda=1$,
standardised features) predicts $t$ from it, with four cross-validation folds split by map so that
no map appears in both training and testing. We report the out-of-fold $R^2$ and three
references: the same network at its initial weights, the raw observation (what is readable
without memory), and the trained state with shuffled labels \citep{hewitt2019control}.

The same analysis was first run on the development test panel of Part~I (the end-of-training
model, averaged over its four evaluations); those results were seen before the confirmatory
analysis was frozen. The analysis code is \texttt{ucl study-contrasts}, checked on synthetic data
with known answers.

\section{Part I: what a memory gap is made of}
\label{sec:results}

All numbers in this section come from the 60 runs of plan \texttt{C08-PARTE1-TALLER} (30 seeds
$\times$ 2 architectures) and their re-evaluations. Unless stated otherwise, intervals are 95\%
percentile bootstrap intervals over seeds (10,000 resamples) and ``pp'' means percentage points.

\subsection{The LSTM outperforms the memory-free MLP on every seed but one}
\label{sec:results-gap}

\begin{figure}[t]
  \centering
  \begin{minipage}[t]{0.49\linewidth}
    \centering
    \includegraphics[width=\linewidth]{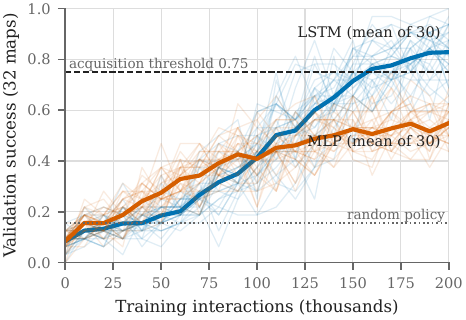}
    \caption{Validation success during training (one evaluation per point, 32 maps). Thin
    lines are individual seeds; thick lines are means over 30 seeds.}
    \label{fig:curves}
  \end{minipage}\hfill
  \begin{minipage}[t]{0.49\linewidth}
    \centering
    \includegraphics[width=\linewidth]{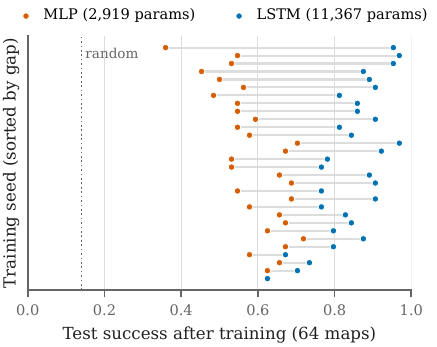}
    \caption{Final test success of each training seed for both architectures, sorted by the
    paired gap. The LSTM is ahead on 29 seeds and tied on one.}
    \label{fig:paired}
  \end{minipage}
\end{figure}

\Cref{fig:curves} shows the learning curves. Both agents start near the random policy. The MLP
improves steadily but levels off around 50--55\% validation success; the LSTM overtakes it after
about 100,000 interactions and keeps improving to above 80\%. On the test panel
(\cref{tab:summary}) the LSTM reaches 84.0\% final success (SD across seeds 8.7\pp) and the MLP
58.9\% (SD 8.2\pp). The paired gap is $+25.1$\pp{} $[20.8, 29.5]$; the LSTM is ahead on 29 of the
30 seeds and tied on the remaining one (\cref{fig:paired}). None of 200,000 random sign flips of
the paired gaps produced a mean as extreme as the observed one ($p<10^{-5}$). The validation AUC
tells the same story with a smaller gap ($+5.8$\pp{} $[4.0, 7.6]$), because both curves are low
during the first half of training.

By the protocol's acquisition rule, 26 of 30 LSTM runs acquire task~B, at a median of 170,000
interactions (range 130,000--190,000); no MLP run does. Part~I therefore confirms the pilot's
observation with ten times more seeds. It does not say why the LSTM wins: it is also 3.9 times
larger. \Cref{sec:part2} addresses that.

\begin{table}[t]
  \centering
  \caption{Part I summary over 30 paired seeds. Final test success is one evaluation on 64 test
  maps; ``re-evaluated'' averages three further evaluations with new sampling seeds. Brackets
  are 95\% bootstrap intervals over seeds; IQM is the interquartile mean
  \citep{agarwal2021precipice}.}
  \label{tab:summary}
  \small
  \begin{tabular}{lccc}
    \toprule
    & LSTM & MLP & LSTM $-$ MLP (paired) \\
    \midrule
    Trainable parameters & 11,367 & 2,919 & \\
    Final test success (mean) & 84.0 [80.8, 86.9] & 58.9 [55.9, 61.7] & $+25.1$ [20.8, 29.5] \\
    Final test success (IQM) & 84.9 & 59.2 & $+24.3$ \\
    Final test success, re-evaluated & 83.9 [81.0, 86.7] & 58.7 [56.5, 61.0] & \\
    Validation AUC & 44.6 [43.0, 46.1] & 38.8 [37.9, 39.7] & $+5.8$ [4.0, 7.6] \\
    Seeds acquiring B & 26 / 30 & 0 / 30 & \\
    Median acquisition step & 170k & --- & \\
    Test success before training & 11.3 & 12.2 & \\
    \bottomrule
  \end{tabular}
\end{table}

\subsection{Runs are exactly reproducible, and one evaluation is unbiased}
\label{sec:results-repro}

The pilot seeds (7, 19, 31) were run again inside Part~I, now saving a checkpoint at every
evaluation. For all six of them (both architectures) the final weights are identical, bit for
bit, to the pilot's, and so are their test scores. Saving extra checkpoints and re-evaluating
them did not perturb training.

Re-evaluating the final weights three more times leaves the means unchanged (84.0 versus 83.9 for
the LSTM, 58.9 versus 58.7 for the MLP; \cref{tab:summary}): the single evaluation is not biased.
It is, however, noisy. Across seeds, the scores of two re-evaluations of the same weights
correlate at only 0.81 for the LSTM and 0.70 for the MLP.

\subsection{Where the variation comes from}
\label{sec:results-variance}

\begin{figure}[t]
  \centering
  \includegraphics[width=\linewidth]{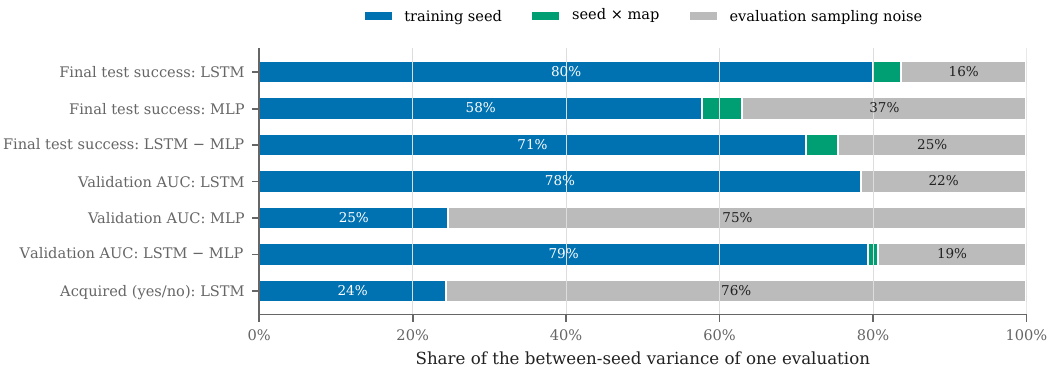}
  \caption{Composition of the between-seed variance of a run's score from one evaluation, for
  each metric. Grey is the part due only to the randomness of evaluation. The MLP never
  acquires task B, so its acquisition verdict has no variance and is omitted.}
  \label{fig:composition}
\end{figure}

\Cref{tab:components} gives the variance components of \cref{eq:model} for final test success,
and \cref{fig:composition} turns them into the quantity that matters for a comparison: of the
differences we see between the scores of different seeds, how much is real (seed, and seed
$\times$ map on this panel) and how much is evaluation noise.

For final test success the answer is reassuring. Evaluation noise explains 16\% of the
between-seed variance for the LSTM, 37\% for the MLP and 25\% for the paired gap; most of what
separates seeds is real. Two other metrics behave very differently:
\begin{itemize}
  \item \textbf{Validation AUC of the MLP} is 75\% noise. MLP seeds learn almost identically
  ($\hat\sigma^2_{\text{seed}}\approx 0.0002$), so the little variation there is between them is
  mostly the randomness of 32-map evaluations. The AUC of the \emph{gap}, by contrast, is only
  19\% noise.
  \item \textbf{The acquisition verdict} is 76\% noise. Whether a given LSTM run ``acquired'' the
  task depends mostly on the sampling seeds of its validation episodes: replaying the same
  checkpoints with the three new sampling seeds changes the verdict for 8 of the 30 LSTM runs.
  The component interval is wide ($\sigma^2_{\text{seed}}\in[0, 0.082]$) and includes zero.
\end{itemize}
The map is the largest single component for final test success (\cref{tab:components}). It does
not add uncertainty to a comparison on the fixed test panel, which both architectures share, but
it dominates any claim about maps outside the panel (\cref{sec:results-seeds}).

\begin{table}[t]
  \centering
  \caption{Variance components of final test success per episode (\cref{eq:model}; $S=30$,
  $M=64$, $R=3$), with 95\% seed-bootstrap intervals.}
  \label{tab:components}
  \small
  \setlength{\tabcolsep}{4pt}
  \begin{tabular}{lccc}
    \toprule
    Component & LSTM & MLP & LSTM $-$ MLP \\
    \midrule
    $\sigma^2_{\text{seed}}$ & 0.0059 [0.0030, 0.0087] & 0.0032 [0.0012, 0.0053] & 0.0096 [0.0043, 0.0151] \\
    $\sigma^2_{\text{map}}$ & 0.0346 [0.0275, 0.0469] & 0.0911 [0.0820, 0.1056] & 0.0368 [0.0332, 0.0486] \\
    $\sigma^2_{\text{seed}\times\text{map}}$ & 0.0174 [0.0104, 0.0210] & 0.0184 [0.0090, 0.0231] & 0.0362 [0.0191, 0.0461] \\
    $\sigma^2_{\text{noise}}$ & 0.0778 [0.0661, 0.0906] & 0.1313 [0.1210, 0.1413] & 0.2111 [0.1918, 0.2332] \\
    \midrule
    Noise share of one evaluation & 16\% & 37\% & 25\% \\
    \bottomrule
  \end{tabular}
\end{table}

\subsection{How many seeds a claim needs}
\label{sec:results-seeds}

\begin{figure}[t]
  \centering
  \begin{minipage}[c]{0.49\linewidth}
    \centering
    \includegraphics[width=\linewidth]{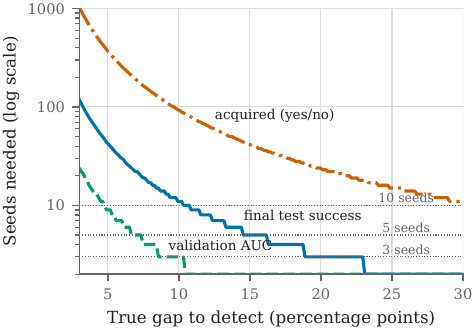}
  \end{minipage}\hfill
  \begin{minipage}[c]{0.47\linewidth}
    \caption{Seeds needed for a paired two-sided test ($\alpha=0.05$, power 0.8) to detect a
    true gap between LSTM and MLP, with one evaluation per run, by metric. Dotted lines mark
    common seed counts. For the acquisition verdict the gap is a difference in the fraction of
    seeds that acquire the task.}
    \label{fig:seeds}
  \end{minipage}
\end{figure}

\Cref{fig:seeds} and \cref{tab:seeds} apply \cref{eq:dstudy} to the paired differences. With one
evaluation per run, as is common practice:
\begin{itemize}
  \item \textbf{Final test success.} Detecting a true gap of 20\pp{} takes 3 seeds, 10\pp{} takes
  11, and 5\pp{} takes 43. The pilot's three seeds could only have supported a gap of about
  20\pp{} or more. Three re-evaluations per run lower these numbers only slightly (43 to 36 for
  5\pp), because most of the per-seed variance is real seed variation, which more evaluations
  cannot remove.
  \item \textbf{Validation AUC} needs the fewest seeds (9 for 5\pp, 3 for 10\pp): averaging over
  the whole curve removes much of the noise. It measures speed and level together, however, so a
  gap in AUC is not a gap in final success.
  \item \textbf{The acquisition verdict} needs 93 seeds to detect a 10\pp{} difference in the
  fraction of seeds that acquire, and 369 for 5\pp. Here more evaluations do help (93 to 46 with
  three re-evaluations), because the verdict is mostly noise.
\end{itemize}
Claims about new maps rather than the fixed panel add the map term of \cref{eq:dstudy}. With 64
test maps that term alone is larger than a $\pm5$\pp{} target for the MLP, so no number of seeds
reaches that precision; for the paired gap the same target needs 178 seeds instead of 21. More
test maps, not more seeds, are what such claims require.

\begin{table}[t]
  \centering
  \caption{Seeds needed for the paired LSTM $-$ MLP comparison on the fixed test panel, with
  $n_r$ evaluations per run. Left: to detect a true gap $\delta$ ($\alpha=0.05$, power 0.8).
  Right: for a 95\% interval of half-width 5\pp{} around the mean gap.}
  \label{tab:seeds}
  \small
  \begin{tabular}{lcccccccc}
    \toprule
    & \multicolumn{2}{c}{$\delta=5$\pp} & \multicolumn{2}{c}{$\delta=10$\pp} &
      \multicolumn{2}{c}{$\delta=20$\pp} & \multicolumn{2}{c}{$\pm5$\pp{} interval} \\
    \cmidrule(lr){2-3}\cmidrule(lr){4-5}\cmidrule(lr){6-7}\cmidrule(lr){8-9}
    Metric & $n_r=1$ & $n_r=3$ & $n_r=1$ & $n_r=3$ & $n_r=1$ & $n_r=3$ & $n_r=1$ & $n_r=3$ \\
    \midrule
    Final test success & 43 & 36 & 11 & 9 & 3 & 3 & 21 & 18 \\
    Validation AUC & 9 & 8 & 3 & 2 & 2 & 2 & 5 & 4 \\
    Acquired (yes/no) & 369 & 183 & 93 & 46 & 24 & 12 & 181 & 90 \\
    \bottomrule
  \end{tabular}
\end{table}

\paragraph{Summary of Part I.}
On this task the LSTM's advantage over a memory-free MLP is large and robust: 25\pp, on 29 of 30
seeds. Measuring it reliably is easy for final success and AUC, and hard for the binary
acquisition verdict that learning curves are often summarised with. Most of the variation between
seeds in final success is real, so it is reduced by more seeds rather than more evaluations, and
claims beyond the test panel are limited by the number of maps.

\section{Part II: the memory advantage is mostly a clock}
\label{sec:part2}

\subsection{The arms on the sealed panel}
\label{sec:part2-arms}

\begin{figure}[t]
  \centering
  \begin{minipage}[t]{0.49\linewidth}
    \centering
    \includegraphics[width=\linewidth]{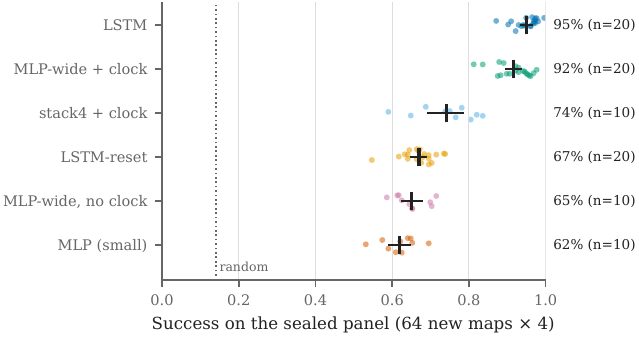}
    \caption{Sealed-panel success of every seed (dots), with the mean and its 95\% bootstrap
    interval (black).}
    \label{fig:part2-arms}
  \end{minipage}\hfill
  \begin{minipage}[t]{0.49\linewidth}
    \centering
    \includegraphics[width=\linewidth]{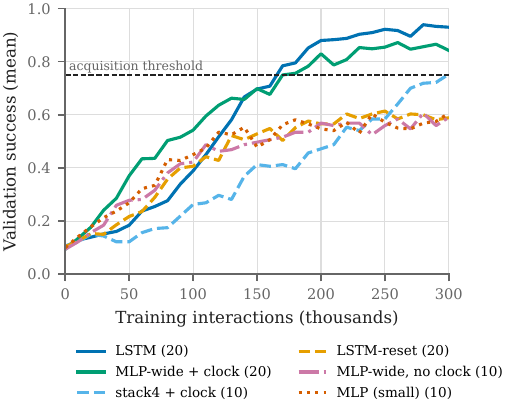}
    \caption{Mean validation success during training (number of seeds in brackets).}
    \label{fig:part2-curves}
  \end{minipage}
\end{figure}

\Cref{tab:arms} and \cref{fig:part2-arms} give the confirmatory score of every arm. The LSTM
succeeds on 95.0\% of the sealed episodes. A memory-free MLP of the same size that is told the
time, MLP-wide + clock, reaches 91.7\%, and like the LSTM it acquires the task on every seed. The
same network without the clock falls to 65.1\%, close to the small MLP (61.9\%) and to the LSTM
whose memory is erased at every step (67.0\%). stack4 + clock is in between (74.2\%), with a wide
spread across seeds. \Cref{fig:part2-curves} shows that MLP-wide + clock even learns faster than
the LSTM during the first half of training, and that the three arms without a way to tell time
level off together around 55--60\% validation success.

\begin{table}[t]
  \centering
  \caption{Part II arms. Sealed: success of the validation-selected checkpoint on 64 new maps
  $\times$ 4 sampling seeds, mean over seeds with a 95\% bootstrap interval. Acquired: seeds
  meeting the acquisition rule within 300,000 interactions, with the median step among them.
  AUC: normalised area under the validation curve.}
  \label{tab:arms}
  \small
  \begin{tabular}{lrcccccc}
    \toprule
    Arm & Params & Clock & lr & Seeds & Sealed (\%) & Acquired & AUC \\
    \midrule
    LSTM & 11,367 & no & $10^{-3}$ & 20 & 95.0 [93.7, 96.3] & 20/20 (175k) & 0.60 \\
    MLP-wide + clock & 11,399 & yes & $10^{-3}$ & 20 & 91.7 [89.7, 93.5] & 20/20 (180k) & 0.62 \\
    stack4 + clock & 11,381 & yes & $3\cdot10^{-4}$ & 10 & 74.2 [69.3, 78.5] & 5/10 (270k) & 0.38 \\
    LSTM-reset & 11,367 & no & $10^{-3}$ & 20 & 67.0 [65.1, 68.8] & 2/20 (240k) & 0.45 \\
    MLP-wide, no clock$^\dagger$ & 11,401 & no & $10^{-3}$ & 10 & 65.1 [62.6, 67.6] & 0/10 & 0.44 \\
    MLP (small) & 2,919 & no & $10^{-3}$ & 10 & 61.9 [59.1, 64.6] & 0/10 & 0.46 \\
    \midrule
    LSTM-reset + clock$^\ddagger$ & 11,495 & yes & $10^{-3}$ & 10 & 91.5 [89.3, 93.4] & 10/10 (120k) & 0.68 \\
    MLP (small) + clock$^\ddagger$ & 2,951 & yes & $10^{-3}$ & 10 & 90.7 [88.4, 92.8] & 10/10 (205k) & 0.60 \\
    \bottomrule
    \multicolumn{8}{l}{\footnotesize $^\dagger$Added after an interim look; exploratory.
    $^\ddagger$Added after the confirmatory result; exploratory.}
  \end{tabular}
\end{table}

\subsection{Confirmatory contrasts}
\label{sec:part2-contrasts}

\Cref{tab:contrasts} gives the preregistered contrasts on the sealed panel.

\textbf{P1 is not supported.} The LSTM is ahead of MLP-wide + clock by only 3.3\pp{}
$[0.9, 6.0]$, on 12 of 20 seeds. The difference is unlikely to be zero (Holm-adjusted
$p=0.041$), but it is far below the 15\pp{} the design required, and the AUC goes the other way
(on the development panel, $-2.8$\pp{} $[-4.6, -0.4]$). The 25 to 34\pp{} advantage of the LSTM
over the small MLP in the pilot and in Part~I is therefore not, for the most part, an advantage of
memory.

\textbf{P2 is supported.} Erasing the LSTM's state at every step costs 28.0\pp{} $[25.3, 29.9]$,
on all 20 seeds: the LSTM's advantage does depend on carrying a state from one step to the next.

\textbf{H3 is supported.} stack4 + clock stays well below the average of the LSTM and
MLP-wide + clock; it is even 17.5\pp{} below MLP-wide + clock, which has the same clock but no
stacked frames.

\begin{figure}[t]
  \centering
  \begin{minipage}[c]{0.52\linewidth}
    \centering
    \includegraphics[width=\linewidth]{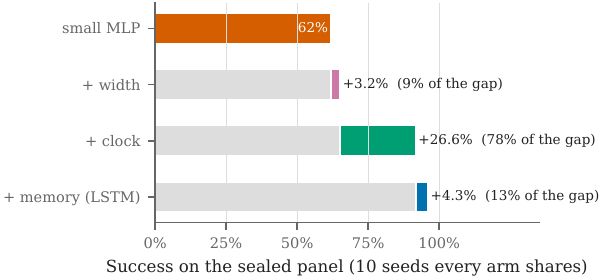}
  \end{minipage}\hfill
  \begin{minipage}[c]{0.44\linewidth}
    \caption{From the small MLP to the LSTM, one ingredient at a time, on the 10 seeds that every
    arm shares (sealed panel). Width alone adds little; telling a memory-free network the time
    closes most of the gap.}
    \label{fig:part2-ladder}
  \end{minipage}
\end{figure}

P1 and P2 together say that the LSTM needs its recurrent state, but that a network with no
memory at all does almost as well if it is told the time. The exploratory contrasts locate the
difference. Adding the clock to the wide MLP adds 26.6\pp{} $[21.5, 30.5]$ on all 10 shared
seeds; adding width without the clock adds 3.2\pp{} $[-0.4, 5.8]$, an interval that includes
zero. \Cref{fig:part2-ladder} adds the ingredients one at a time on the 10 seeds every arm shares:
of the 34\pp{} between the small MLP and the LSTM, width accounts for 9\%, the clock for 78\% and
what the LSTM does beyond a clock for 13\%.

The development panel gives the same verdicts: P1 is $+4.9$\pp{} $[2.3, 8.0]$ and not
supported, P2 is $+32.1$\pp{} and H3 is supported. The conclusion therefore does not depend on
which panel or checkpoint is used.

\begin{table}[t]
  \centering
  \caption{Contrasts on the sealed panel (percentage points). $p$: exact two-sided sign-flip
  test; Holm: adjusted within the family ($m=3$ principal, $m=5$ secondary). Ahead: seeds on
  which the contrast is positive. The exploratory family has no correction and no verdict.}
  \label{tab:contrasts}
  \footnotesize
  \setlength{\tabcolsep}{4pt}
  \begin{tabular}{llrcrrrc}
    \toprule
    & Contrast & Mean & BCa 95\% & $p$ & Holm & Ahead & Verdict \\
    \midrule
    P1 & LSTM $-$ MLP-wide + clock & $+3.3$ & [0.9, 6.0] & 0.021 & 0.041 & 12/20 & not supported$^*$ \\
    P2 & LSTM $-$ LSTM-reset & $+28.0$ & [25.3, 29.9] & $2\cdot10^{-6}$ & $6\cdot10^{-6}$ & 20/20 & supported \\
    H3 & LSTM $+$ MLP-wide $-$ 2 stack4 & $+39.2$ & [32.5, 47.9] & 0.002 & 0.010 & 10/10 & supported \\
    \midrule
    E1 & clock: MLP-wide $\pm$ clock & $+26.6$ & [21.5, 30.5] & 0.002 & & 10/10 & \\
    E2 & width: MLP-wide, no clock $-$ MLP & $+3.2$ & [$-0.4$, 5.8] & 0.092 & & 9/10 & \\
    E3 & LSTM $-$ MLP (small) & $+34.1$ & [31.6, 38.2] & 0.002 & & 10/10 & \\
    E4 & LSTM $-$ stack4 & $+21.8$ & [17.8, 27.4] & 0.002 & & 10/10 & \\
    E5 & stack4 $-$ MLP-wide + clock & $-17.5$ & [$-21.0$, $-13.6$] & 0.002 & & 0/10 & \\
    E6 & LSTM-reset $-$ MLP (small) & $+5.5$ & [2.1, 9.2] & 0.018 & & 8/10 & \\
    \bottomrule
  \end{tabular}

  \smallskip
  {\footnotesize $^*$Holm $p<0.05$ and interval above 0, but the mean is below the 15\,pp
  threshold.}
\end{table}

\subsection{Why a clock?}
\label{sec:part2-why}

The task suggests a reason. Episodes are cut at 32 steps and the
success reward, $1-0.9\,t/32$, falls with time, so the value of a state depends on how many steps
are left. A critic that cannot tell time sees the same observation early and late in an episode
and must average two different values, which degrades the advantage estimates that drive A2C's
policy update; this is the failure mode \citet{pardo2018timelimits} describe. An LSTM can count
steps in its state, and LSTM-reset cannot, which would explain why P2 is large while P1 is small.

\paragraph{The LSTM's state encodes the time step.}
A linear probe reads the step $t$ from the trained LSTM's state with an out-of-fold $R^2$ of 0.84
(mean over 20 seeds; \cref{tab:probe}). Some of this is free: at its initial weights the same
network already reaches 0.52, because a recurrent state drifts as it integrates inputs. Training
increases it on 19 of 20 seeds. The observation alone carries no usable information about $t$
($R^2<0$), nor does LSTM-reset, whose state is erased at every step. The LSTM therefore has, and
learns to sharpen, exactly the information the clock gives the memory-free arms.

\paragraph{The time step alone closes the gap, whatever the network.}
A probe shows that the information is there, not that it matters. The two mechanism arms test
that directly (\cref{tab:arms,tab:mechanism}). Giving the clock to LSTM-reset, the network whose
memory is erased at every step, raises it from 67.0\% to 91.5\% (+24.1\pp{} $[21.6, 27.0]$ on all
10 seeds): of the 28 points the LSTM loses when its state is erased, the clock alone gives back
24. LSTM-reset + clock also learns faster than the LSTM (it acquires the task at a median of
120,000 interactions, against 175,000). Giving the clock to the small MLP, a quarter of the
LSTM's size, raises it from 61.9\% to 90.7\% (+28.8\pp), and once the clock is there width adds
nothing measurable (MLP-wide + clock minus MLP + clock, $+1.0$\pp{} $[-2.7, 5.1]$). What remains
for the LSTM over a memory-free network that knows the time is 4.5 to 5.3\pp, in line with P1.
On this task, a recurrent state's main contribution is to know the time; a single input does the
same job.

\begin{table}[t]
  \centering
  \caption{Mechanism contrasts on the sealed panel (percentage points; exploratory, 10 paired
  seeds, exact sign-flip $p$, no correction).}
  \label{tab:mechanism}
  \small
  \begin{tabular}{llrcrr}
    \toprule
    & Contrast & Mean & BCa 95\% & $p$ & Ahead \\
    \midrule
    M1 & LSTM-reset + clock $-$ LSTM-reset & $+24.1$ & [21.6, 27.0] & 0.002 & 10/10 \\
    M2 & LSTM $-$ LSTM-reset + clock & $+4.5$ & [2.5, 6.9] & 0.002 & 10/10 \\
    M3 & MLP + clock $-$ MLP & $+28.8$ & [24.0, 31.6] & 0.002 & 10/10 \\
    M4 & MLP-wide + clock $-$ MLP + clock & $+1.0$ & [$-2.7$, 5.1] & 0.66 & 6/10 \\
    M5 & LSTM $-$ MLP + clock & $+5.3$ & [2.6, 8.6] & 0.012 & 8/10 \\
    \bottomrule
  \end{tabular}
\end{table}

\begin{table}[t]
  \centering
  \caption{Linear probe for the episode step $t$ (out-of-fold $R^2$, folds split by map; mean
  over 20 seeds per arm). Episodes last 9.5 steps on average because successful ones end early.}
  \label{tab:probe}
  \small
  \begin{tabular}{lcccc}
    \toprule
    Features & Trained state & Initial-weights state & Observation & Shuffled labels \\
    \midrule
    LSTM & 0.84 & 0.52 & $-0.19$ & $-0.10$ \\
    LSTM-reset & 0.04 & $-0.07$ & $-0.06$ & $-0.03$ \\
    \bottomrule
  \end{tabular}
\end{table}

\paragraph{Summary of Part II.}
On task B, a memory-free network that is told the time comes within 3 to 5\pp{} of the LSTM on
new maps, whether it is an MLP of the LSTM's size, an MLP a quarter of that size, or the LSTM itself
with its memory erased. Most of what looked like a memory advantage is an advantage of knowing how
much time is left; the part that is specific to memory is small, though probably not zero.

\section{Discussion}
\label{sec:discussion}

\paragraph{What the memory advantage was.}
The LSTM's 25 to 34-point advantage over a small memory-free MLP survives every statistical check
we applied, and it is real in the sense that it reproduces across seeds and maps. It is not,
for the most part, an advantage of remembering what was seen. Width alone explains about a tenth
of it, the time step about four fifths, and what is left for the LSTM beyond a clock is about
3 points on new maps, a difference we can detect but that is far smaller than the one we set out
to explain. The LSTM does need its recurrent state (erasing it costs 28 points), and its state
does encode the time step, and handing the time step to the same network with its state erased
gives back 24 of those 28 points. On this task, the recurrent state is used largely as a clock. Whether the remaining 3 points are memory of the scene or a
better clock is an open question.

\paragraph{Recommendations.}
For memory comparisons in tasks with time limits or time-dependent rewards:
\begin{enumerate}
  \item Give memory-free baselines the time step, or remove the dependence on time, before
  attributing a gap to memory. In our task this one input moved a memory-free network from 65\%
  to 92\%.
  \item Match parameters, and include an ablation that keeps the recurrent agent's weights but
  erases its state; together they separate capacity from recurrence.
  \item Report final success over paired seeds, with intervals: about ten paired seeds detect a
  10-point gap here, and about forty a 5-point gap. Avoid binary acquisition verdicts, or report
  them with many seeds and several evaluations per checkpoint.
  \item Re-evaluate saved checkpoints with new sampling seeds. It is cheap, it measures the noise
  of the evaluation directly, and it showed that the acquisition verdict of more than a quarter of
our LSTM runs (8 of 30) flips under new sampling seeds.
  \item Keep a sealed panel of new maps for the final comparison, and use more maps rather than
  more seeds when the claim is about generalization to new maps.
\end{enumerate}

\paragraph{Variance in procedurally generated environments.}
The decomposition shows why a single number such as ``use ten seeds'' cannot be right in general.
Final success varied mainly between seeds, so more seeds help and more evaluations help little;
the acquisition verdict varied mainly between evaluations, so the opposite holds; and claims about
unseen maps are limited by the number of maps, whatever the number of seeds. A short pilot with
re-evaluation is enough to estimate these components and to choose a design before the main
experiment.

\paragraph{Scope.}
Everything here is one task, one algorithm and small networks, and we expect the size of the
clock effect to depend on the horizon, on how strongly the reward depends on time and on how
value targets are computed at the time limit. The method, however, transfers: the same controls
and the same decomposition can be run on any memory benchmark at little cost, and we release the
tooling to do so.

\section{Limitations}
\label{sec:limitations}
All runs use one task (B), one environment family, one algorithm (A2C) and small networks
(2,919 and 11,367 parameters) on a $7\times7$ grid. The variance components are specific to
this setting; the method that estimates them is not. The test panel is fixed, so claims about
new maps carry an additional map term (\cref{sec:results-seeds}). The acquisition verdict uses
the protocol's rule as written; other thresholds would change its noise share.

In Part~II, the hypotheses and thresholds were fixed in a written design before any run, but not
registered on a public platform. P3, the fully observable twin, was not run. The clock control
was added after an interim look and is exploratory, and its 10 seeds give wide intervals for the
width effect; the mechanism arms were added after the confirmatory result, also on 10 seeds,
and the sealed panel had already been opened once when they were played on it (it was still
unseen by them). The account of why the clock matters (\cref{sec:part2-why}) rests on a probe and
on these arms; we did not test it with a reward that ignores time. Whether the result holds for other tasks, longer horizons, rewards that do not depend on
time, or other algorithms is open.

\section{Conclusion}
\label{sec:conclusion}
A clear memory advantage on a partially observable XLand-MiniGrid task turned out to be mostly
an advantage of knowing the time. We found this by separating the noise of the measurement from
real differences between agents and by controlling, one at a time, for capacity, recurrence,
short-term memory and time awareness, on maps no agent had seen. Claims that an architecture
benefits from memory should survive the same controls.

\section*{Reproducibility statement}
Every run was listed in a plan file before any of its own results existed, executed by an
event-driven
chain that records each step in an append-only journal, and re-evaluated from saved weights.
Re-running a seed reproduces its final weights bit for bit (\cref{sec:results-repro}). Code,
plans, configurations and analysis specifications are available at \codeurl; per-episode
outcomes and all checkpoints are archived at \dataurl.

\bibliographystyle{plainnat}
\bibliography{references}

\end{document}